\documentclass[runningheads]{llncs}

\usepackage{amssymb}
\usepackage[T1]{fontenc}
\usepackage{amsmath}
\usepackage{graphicx}
\usepackage{float}
\usepackage{amsfonts}
\usepackage{booktabs}
\usepackage{epsfig}
\usepackage{subfigure}
\usepackage{multirow}
\usepackage{xcolor}
\usepackage{url}
\usepackage[colorlinks=true,allcolors=blue]{hyperref}
\title{Mending of Spatio-Temporal Dependencies in Block Adjacency Matrix}

\author{Osama Ahmad\inst{1} \and
Omer Abdul Jalil \inst{1}\and
 Usman Nazir \inst{2} \and
Murtaza Taj \inst{1}
 }
\authorrunning{O. Ahmad et al.}
\institute{Lahore University of Management Sciences \and
Beaconhouse National University\\
\email{\{22060007, 23100050, murtaza.taj\}@lums.edu.pk}\\
\email{usman.nazir@bnu.edu.pk}
}

\begin{document}

\maketitle

\begin{abstract}
In the realm of applications where data dynamically evolves across spatial and temporal dimensions, Graph Neural Networks (GNNs) are often complemented by sequence modeling architectures, such as RNNs and transformers, to effectively model temporal changes. These hybrid models typically arrange the spatial and temporal learning components in series. A pioneering effort to jointly model the spatio-temporal dependencies using only GNNs was the introduction of the Block Adjacency Matrix \(\mathbf{A_B}\) \cite{1}, which was constructed by diagonally concatenating adjacency matrices from graphs at different time steps. This approach resulted in a single graph encompassing complete spatio-temporal data; however, the graphs from different time steps remained disconnected, limiting GNN message-passing to spatially connected nodes only. Addressing this critical challenge, we propose a novel end-to-end learning architecture specifically designed to mend the temporal dependencies, resulting in a well-connected graph. Thus, we provide a framework for the learnable representation of spatio-temporal data as graphs. Our methodology demonstrates superior performance on benchmark datasets, such as SurgVisDom and C2D2, surpassing existing state-of-the-art graph models in terms of accuracy. Our model also achieves significantly lower computational complexity, having far fewer parameters than methods reliant on CLIP and 3D CNN architectures.

\end{abstract}
\section{Introduction}
\label{sec:intro}
The arena of spatio-temporal data analysis has rapidly expanded within the domain of deep learning, finding applications in diverse sectors such as traffic management~\cite{traffic_forecasting}, agricultural yield prediction~\cite{yield_pred}, flood risk assessment~\cite{flood_forecast}, land cover monitoring~\cite{land_cover}, ride-sharing demand prediction~\cite{ride_hailing}, video-based activity recognition~\cite{video_action}, and brain-computer interfaces like EEG-based intention recognition~\cite{eeg}. Spatio-temporal data is inherently complex due to its reliance on both spatial configurations and temporal evolution~\cite{ST_data}, necessitating the development of sophisticated models that can handle both these dimensions effectively.

 In complex systems that contain both spatial and temporal dependencies (\textbf{spatio-temporal dependencies}), the existing deep learning methods use two separate networks, temporal learning network and spatial learning network~\cite{STGNN} to separately model the temporal and spatial relations.  To determine the temporal variation, ~\cite{splitting_algorithm} formulates a convex optimization method assuming that the dynamic changes in consecutive events are sparse, this approach applies to a small number of observations. For large complex datasets, the temporal learning network is typically a Recurrent Neural Network (RNN) \cite{RNN} or a Temporal Convolution Network (TCN) \cite{TCN}, and the spatial learning network is typically a Graph Convolution Network (GCN) \cite{GCNconv} or a Graph Attention Network (GAT) \cite{Graph_attention_networks}. The resulting features from both these branches are then concatenated or fused using a spatio-temporal fusion neural architecture and used for the downstream tasks of interest. 

Graph Neural Networks (GNNs) have shown promise in modeling spatial data that is graph-based, but they lack the inherent capability to process sequences. They need to be augmented by Recurrent Neural Networks (RNNs)~\cite{euler} or transformers~\cite{modular_graph_transformer,STAR,GTN} for modeling the temporal dependencies. A notable attempt to overcome the limitations of GNN was the introduction of the Block Adjacency Matrix~\cite{1}. Block adjacency aimed to encapsulate dynamic spatio-temporal relationships within a single graph by concatenating the adjacency matrices of all time steps into a single matrix. This allowed a single GNN to be used for joint spatio-temporal representation learning. However, the approach failed to establish connections between graphs from different time steps, leading to isolated sub-graphs and consequently limiting the message-passing capabilities of GCNs.  

Our work introduces an encoder block to infer the missing temporal connections, thereby providing a learnable framework for the joint graph that captures the full spatio-temporal topology. Our key contributions are as follows:
\begin{enumerate}
    \item A transformer-based encoding architecture that mends block adjacency by introducing connections among graphs from different time-steps thus providing a learnable framework for the structure of spatio-temporal graphs.
    \item To ensure meaningful temporal connections and avoid any unconnected sub-graphs, we propose a loss function that promotes non-zero eigenvalues for the Laplacian of the resulting adjacency matrix by modulating its sparsity denoted by $\rho$.
    \item Our proposed approach incorporates the encoder to improve the connectivity in the graph, the spatial and temporal representation learning is accomplished simultaneously using a GNN.
\end{enumerate}

The rest of this paper is organized as follows: Section~\ref{methodology} describes the problem formulation and framework details. Section~\ref{result} discusses the datasets used and results, including the ablative studies based on the different mending techniques and the sparsity of the modified block adjacency matrix. Finally, section~\ref{conclusion} provides a conclusion and discusses future work. 


\section{Methodology}
\label{methodology}
\subsection{Problem Formulation}
A dynamically evolving graph can be defined as a time-ordered sequence of graphs $[ \ \mathcal{G}_{1}, \mathcal{G}_{2}, \ldots, \mathcal{G}_{T} ]$, where T denotes the total number of time-steps in the sequence. At each time-step \( t \), the graph \( \mathcal{G}_{t} \) is characterized by a triplet \( (\mathbf{V_\textit{t}}, \mathbf{A_\textit{t}}, \mathbf{X_\textit{t}}) \) \cite{GNN_survey}. These components are described in detail as follows:

\begin{itemize}
    \item \textbf{Vertices.} \( \mathbf{V_\textit{t}} \) denotes the set of vertices in the graph at time \( t \). The cardinality of this set is denoted by \( n_t \), i.e., \( |\mathbf{V_\textit{t}}| = n_t \).
    \item \textbf{Adjacency Matrix.} \( \mathbf{A_\textit{t}} \) is an \( n_t \times n_t \) matrix that describes the connections between the vertices. 
    \item \textbf{Feature Matrix.} \( \mathbf{X_\textit{t}} \) is an \( n_t \times d_t \) matrix, where \( d_t \) is the dimensionality of the feature space. 
\end{itemize}


Existing methods on dynamic Graph Neural Networks (GNNs)~\cite{feddy,EvolveGCN}, typically allow varying features and edges over time. However, the addition and removal of nodes are not supported in graph convolution neural networks. In our work, we consider graphs where the number of nodes \( n_t \) can vary over time. The feature dimension \( d_t \) remains constant across all time-steps, i.e., \( d_t = d \) $\forall$ \( t \).

\textbf{\textit{Problem Definition.}} Given a block adjacency matrix ($\mathbf{A_B}$) of the supergraph, the objective is to learn a mapping function \( f \) parameterized by \( \Theta \). The function \( f \) modifies \( \mathbf{A_B} \) into \( \mathbf{\hat{A}_B} \) such that the subgraphs within the supergraph have inter-connectivity, which is measured by the eigenvalues of the corresponding Laplacian matrix.

\begin{equation}
  \mathbf{\hat{A}_B} = f(\mathbf{A_B}; \Theta),
\end{equation}
\(\hspace{80pt}\text{s.t.} \quad \left\| \lambda_{\mathbf{\hat{A}_B}} \right\|_0 > \left\| \lambda_{\mathbf{A_B}} \right\|_0\)

Here, \( \left\| \lambda_{\mathbf{\hat{A}_B}} \right\|_0 \) and \( \left\| \lambda_{\mathbf{A_B}} \right\|_0 \) are the $L_0$ norm of the vector of eigenvalues of the Laplacian matrices corresponding to \( \mathbf{\hat{A}_B} \) and \( \mathbf{A_B} \) respectively. The constraint requires that \( \mathbf{\hat{A}_B} \) has more non-zero eigenvalues than \( \mathbf{A_B} \), thereby reducing the number of disconnected sub-graphs in the graph. 
\begin{figure}[t]
    \centering{\includegraphics[width=0.99\textwidth, height=9cm, keepaspectratio]{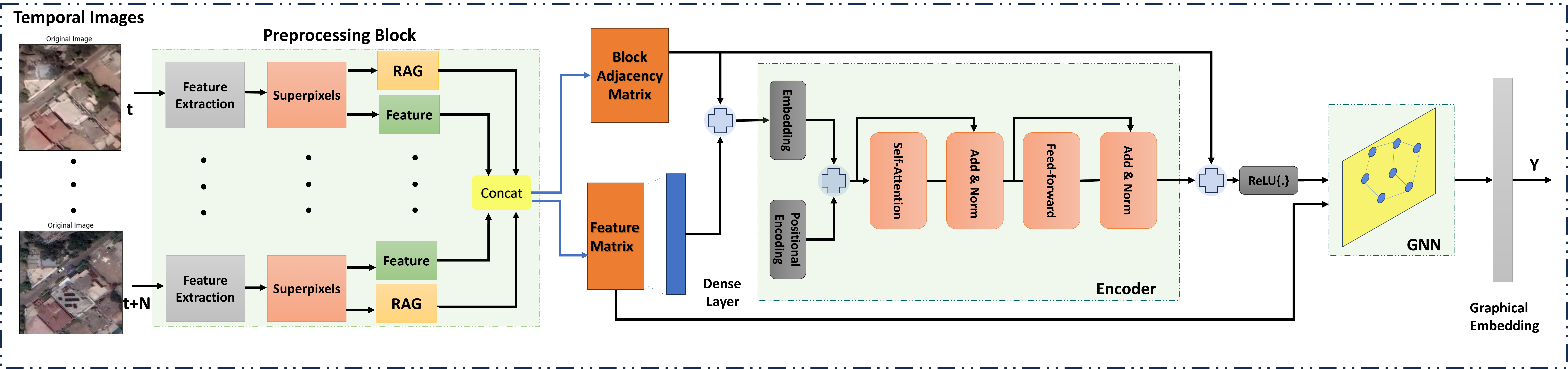}}
    \caption{Proposed Framework: STBAM-GNN  for spatio-temporal analysis consists of 1) Creating a Region Adjacency Graph (RAG) over the super-pixels of the image at each time step, 2) Making a spatio-temporal super-graph by concatenating the adjacency matrices of the graphs at each time step in a Block Adjacency matrix, 3) An encoder for mending BA to capture temporal information, 4) GNN for joint spatio-temporal representation learning, and 5) A task-specific prediction module.}
    \label{fig_1}
\end{figure}
\subsection{Proposed framework}
 An overview of the proposed framework known as spatiotemporal Block Adjacency Matrix (STBAM) is shown in Fig.~\ref{fig_1}. In this work, we focus on spatio-temporal visual data. Our proposed methodology has the following major components:
\begin{enumerate}
   \item Representation of data at each timestep into a Region Adjacency Graph (RAG) $\mathcal{G}_{t}$ using superpixels.
    \item Generating a supergraph of multiple RAGs $\mathcal{G}_t$ through Block Adjacency matrix $\mathbf{A_B}$. 
    \item Learning an encoder for mending $\mathbf{A_B}$ to capture temporal relations.
    \item A GNN to jointly learn the spatio-temporal feature representation of the data from the supergraph.
    \item A task-aware prediction module.
    
\end{enumerate}

The main components of our method Spatio-temporal Block Adjacency Matrix (STBAM) are explained in the following sections.
 \subsubsection{SuperGraph of RAGs}
\begin{figure}[t]
\centering
\includegraphics[width=0.45\textwidth, keepaspectratio]{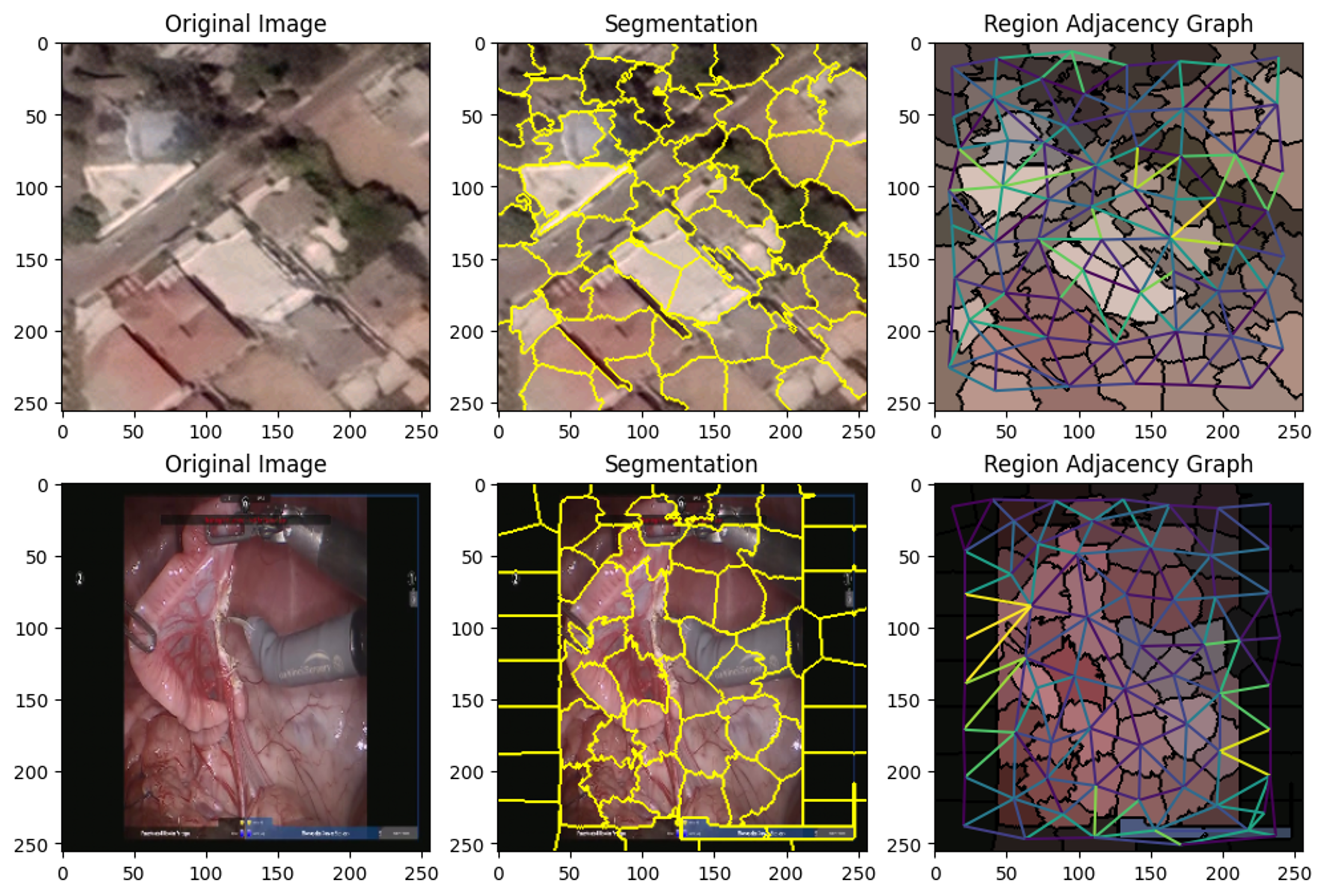}
    \caption{SuperPixel representation on image obtained via SLIC Row 1: Satellite data; Row 2: Surgical data.}
\label{fig:slic}
\end{figure}

 To address the challenges of processing high-dimensional image data as graphs, particularly the significant computational complexities due to the large number of nodes, we adopt graph simplification. This involves the aggregation of nodes with similar features to effectively reduce the size and complexity of the graph. Leveraging superpixels generated via Simple Linear Iterative Clustering (SLIC)~\cite{1,4} (see Fig.~\ref{fig:slic}), we construct a computationally efficiency and manageable memory resources graph representation of images with $N$ superpixels. Here, the number of superpixels $N$ is a hyper-parameter.  In each graph $\mathcal{G}_t$, there are $n_t$ nodes and every node has $d$ features. Features of each node are the sum of the RGB channels, to further enrich node representation we utilize features from Convolutional layer 1 (conv1) of VGG16~\cite{VGG16} trained on ImageNet. To connect the superpixels in RAG, the distance metric is used by SLIC based on color combination and spatial proximity. Finally, these RAGs $\mathcal{G}_t$  are combined into a supergraph by concatenating their adjacency matrices diagonally into a block adjacency matrix $\mathbf{A_B}$, which can be defined as:
\label{1}
\begin{gather}
\mathbf{A_B}=\begin{bmatrix}
\mathbf{A_{1}} & \mathbf{Z} & \cdots & \mathbf{Z} \\
\mathbf{Z} & \mathbf{A_{2}} & \cdots & \mathbf{Z} \\
\vdots & \vdots & \ddots & \vdots \\
\mathbf{Z} & \mathbf{Z} & \cdots & \mathbf{A_T} \\ 
\end{bmatrix},
\end{gather}
where $\mathbf{Z}$ represents matrix of zeros, and $\mathbf{A_t}$ is the adjacency matrix for $\mathcal{G}_t$. To avoid under-segmentation and over-segmentation, we use zero padding and node dilation that provides a consistent number of nodes for images at each timestep. The key characteristic properties of this block adjacency matrix $\mathbf{A_B}$ is that its Laplacian matrix is positive semi-definite (PSD) and contains $n$ zero eigenvalues for $n$ disconnected sub-graphs. The feature representation $\mathcal{X}$ of the supergraph can be arranged as:
\label{eq_2}
\begin{gather}
\mathbf{\mathcal{X}=[X_1,X_2,. . .,X_{T}]^\textit{T}},
\end{gather}

 Where $\mathbf{X}_\textit{t} \in \mathbb{R}^{n \times d}$ represents the node features of the graph at time-step $\textit{t}$. If each node has $d$ features, the order of $\mathcal{X}$ will be  $(n \text{T} \times d)$ where the $n_1$ rows will represent the features for the $n$ nodes of $\mathcal{G}_1$ and so on. The supergraph is thus a stack of sub-graphs, each representing the data at $1$ time step. The supergraph allows a single graph to represent the complete spatio-temporal data, however, these sub-graphs within a supergraph are still disconnected. This limits the message passing within sub-graphs only. Although this methodology allows us to use GNN for spatio-temporal representation learning, the GNN is unable to capture the temporal relation between the data at different time steps. We solve this problem by designing an encoder to mend the missing temporal edges in $\mathbf{A_B}$ as discussed next.
\subsubsection{Encoder for Spatio-temporal Block Adjacency Matrix (STBAM)} 
\begin{figure}[t]
\begin{center}
\includegraphics[width=0.45\textwidth,keepaspectratio]{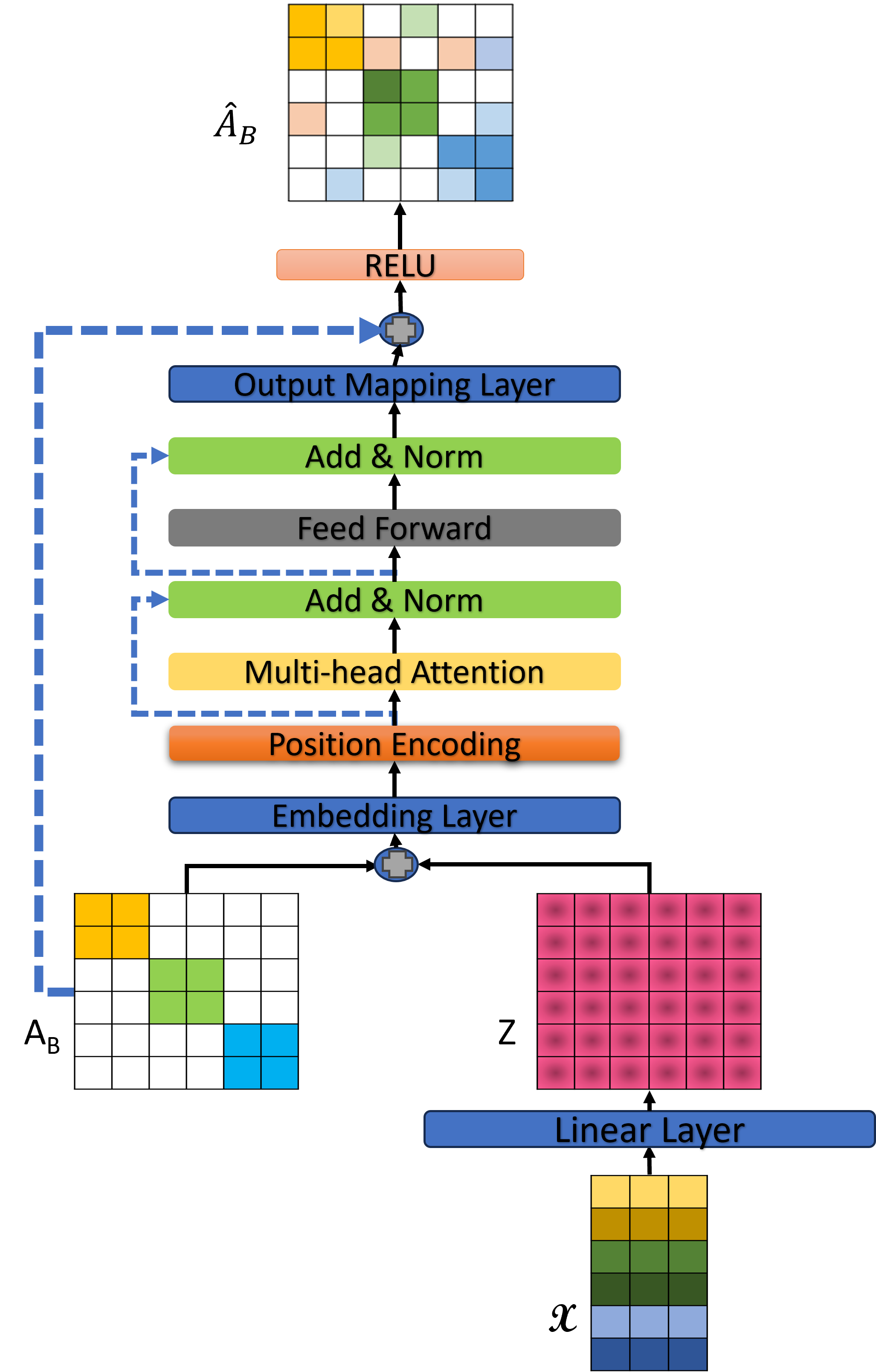}
\end{center}
    \caption{Enhancement of the connectivity using encoding block.}
\label{encoding_block}
\end{figure}
In Fig.~\ref{encoding_block} we show the architecture of the encoder block~\cite{transformer}, which plays a pivotal role in mending the block adjacency matrix $\mathbf{A_B}$. The objective is to create a modified block adjacency matrix $\mathbf{\hat{A}_B}$ that captures the temporal relationships between nodes of the graphs at each time step. The rationale for employing a self-attention based encoder stems from its capability to capture dependencies without regard to their distance in the input data. This attribute is particularly beneficial for learning the missing temporal links between sub-graphs from different time steps represented in a block adjacency matrix. Thus enhancing connectivity and allowing for a more comprehensive representation of spatio-temporal relationships in graphs.

In our approach, a more informative encoding is facilitated by incorporating node features $\mathcal{X}$ into the sparse adjacency matrix. This is achieved by projecting the $\mathcal{X}$ into a higher-dimensional space using a learnable projection layer $\mathbf{W_p}$. The augmented block adjacency matrix $\mathbf{\Tilde{A}_B}$ is formulated as follows:
\begin{equation}
\mathbf{\Tilde{A}_B}=\mathbf{A_B}+(\mathcal{X} \mathbf{W_p}),
\end{equation}
\label{eq_4}

The augmented block adjacency matrix \( \mathbf{\Tilde{A}_B} \) is passed through the encoder. A single layer of the encoder consists of multi-head self-attention and position-wise feed-forward networks, and can be expressed as follows:
\begin{align}
  \text{Attention} = \text{softmax}\left( \frac{\mathbf{W_Q} \mathbf{\Tilde{A}_B} (\mathbf{W_K} \mathbf{\Tilde{A}_B})^{T}}{\sqrt{d}} \right) \mathbf{W_V} \mathbf{\Tilde{A}_B} 
\label{eq:1}, 
\end{align}
here, $\mathbf{W_Q},\mathbf{W_K},\mathbf{W_V}$ are learnable parameters for query, key, and value for attention. The attention score is scaled down by dividing the square root of the embedding dimension $d$.
\begin{align}
\mathbf{\Tilde{A}_B} &= \text{LayerNorm}\left( \mathbf{\Tilde{A}_B} + \text{Attention} \right) \label{eq:3}, \\
\mathbf{\Tilde{A}_{B_{ff}}} &= \left(\text{ReLU}( \mathbf{W_1} \mathbf{\Tilde{A}_{B}} + \mathbf{b_1} )\right) \mathbf{W_2} + \mathbf{b_2} \label{eq:4}, 
\end{align}
$\mathbf{W_1}, \mathbf{W_2}$ are learnable weights, and $\mathbf{b_1}, \mathbf{b_2}$ are bias of the linear layers. ReLU is used to eliminate edges with negative weights. 
\begin{align}
    \mathbf{\hat{A}_{B}} &= \text{LayerNorm}\left( \mathbf{\Tilde{A}_{B}} + \mathbf{\Tilde{A}_{B_{ff}}} \right) \label{eq:5},
\end{align}
The encoder outputs a modified block adjacency matrix $\mathbf{\hat{A}_{B}}$ which is not symmetric and has real values. The matrix is first decomposed into a symmetric matrix by taking the average of the matrix and its transpose, then a skip connection is added to it to reinforce the spatial edges. The final modified adjacency matrix $\mathbf{\hat{A}_{B}}$ can be described as:
\begin{equation}
\mathbf{\hat{A}_{B}} = ReLU\left(\mathbf{A_B}+\frac{\mathbf{\hat{A}_{B}} + \mathbf{\hat{A}_{B}^\textit{T}}} {2}\right),
\end{equation}


\subsubsection{Learning Encoder} 
{\bf Loss function}: Our encoder and GNN are trained end-to-end for spatio-temporal classification using a novel loss function \(\mathcal{L}\), which combines cross-entropy and a sparsity-promoting term:
\begin{equation}
\mathcal{L} = -\sum_{i=1}^{N} y_i \log(\hat{y}_i) + \lambda \lVert \frac{\mathbf{\hat{A}_{B}} - \mathbf{\hat{A}_{B}^\textit{T}}}{2}\rVert_p.
\end{equation}

The hyper-parameter $\lambda$ penalty term must be appropriately adjusted. The sparsity term uses the element-wise L1 norm to encourage a sparse block adjacency matrix thus ensuring only meaningful edges. Training involves the following steps:
\begin{enumerate}
    \item Mend the adjacency matrix \(\mathbf{\hat{A}_{B}} = Encoder(\mathbf{A_{B}})\)
    \item Compute node representations \(\mathbf{z} = GNN(\mathbf{\hat{A}_{B}}, \mathbf{X})\).
    \item Concatenate node representations \(\mathbf{z_{||}} = \text{Concat}(\mathbf{z_1}, \ldots, \mathbf{z_n})\).
    \item Calculate \(\mathbf{h} = \mathbf{W_g}^T \mathbf{z_{||}} + \mathbf{b_G}\).
    \item Obtain output \(\mathbf{y} = \text{softmax}(\mathbf{h})\).
\end{enumerate}

In our approach, we incorporate the Graph Attention Network (GAT) \cite{Graph_attention_networks} as a pivotal component for performing feature transformation and attention-based aggregation. The choice of GAT is motivated by its unique capability to dynamically assign different levels of importance to nodes in a neighborhood, allowing for a more nuanced representation of node interactions. This dynamic weighting mechanism is particularly advantageous in spatio-temporal graph-based learning tasks, where the relevance of temporal and spatial neighbors can vary significantly across different contexts and time steps. By leveraging attention mechanisms, GAT can adapt to the most informative parts of the graph structure, enhancing the model's ability to learn complex dependencies and improving prediction accuracy. The attention in GAT thus serves a different purpose from the attention within the encoder which was solely responsible for introducing edges between the graphs from different time steps. Briefly, the core equations of GAT for layer \( l \) can be summarized as follows:

\begin{align}
    \mathbf{h'}_i^{(l)} &= \mathbf{W}^{(l)} \mathbf{h}_i^{(l)}, \label{eq:gat_linear_transform} \\
    e_{ij}^{(l)} &= \text{LeakyReLU}\left( \mathbf{a}^{(l)T} (\mathbf{h'}_i^{(l)} \Vert \mathbf{h'}_j^{(l)}) \right), \label{eq:gat_attention_coef} \\
    \alpha_{ij}^{(l)} &= \frac{\exp(e_{ij}^{(l)})}{\sum_{k \in \mathcal{N}(i)} \exp(e_{ik}^{(l)})}, \label{eq:gat_alpha} \\
    \mathbf{h}_i^{(l+1)} &= \sigma\left( \sum_{j \in \mathcal{N}(i)} \alpha_{ij}^{(l)} \mathbf{h'}_j^{(l)} \right), \label{eq:gat_output}
\end{align}

Here, \( \mathbf{W}^{(l)} \) is the weight matrix for linear transformation, \( \mathbf{a}^{(l)} \) is the attention mechanism parameter, \( \alpha_{ij}^{(l)} \) is the normalized attention coefficient, and \( \sigma \) is an activation function.

\section{Results}
\label{result}
\subsection{Datasets and Implementation Details}
We perform experiments on two different datasets. 

\textbf{C2D2 dataset:} It is a remote sensing-based spatio-temporal dataset consisting of four classes construction, destruction, cultivation, and de-cultivation~\cite{c2d2}. It contains the temporal samples of the years 2011, 2013, and 2017 from Digital Globe. It visits almost 5,50,000 random locations which make approximately 5310 $km^2$. Where they cropped images of $256 \times 256$ at zoom level 20 which corresponds to 0.149 pixels per meter on the equator. 

\textbf{SurgVisDom dataset:} Surgical Visual Domain Adaptation dataset~\cite{surgvisdom} consists of three class classification tasks: needle-driving (ND), dissection (DS), and knot-tying (KT). The training data consists of 476 total videos, 450 clips from virtual reality, and 26 clips from real-time clinical data. The testing data consists of 16 clips from clinical data. For training we divide each training sample into 16 equal segments and take 1 random frame from each segment, this gives us 16 frames from each sample. We also over-sample every video to increase our training data. We employ a dilated sliding window scheme for frame-by-frame predictions. To predict for frame \( m \), we center the window on frame \( m \) and take 8 frames ahead and 7 frames behind. However, to capture meaningful temporal information—given that actions in surgical tasks unfold gradually—we opt for a dilated window, selecting frames with jumps of 4.

\textbf{Implementation Details:} Experiments are conducted on a Linux machine equipped with an Intel(R) i9 $12^{th}$ Gen @ 2.40 GHz, 32 GB RAM, and NVIDIA 3080Ti GPU. The proposed method was implemented using PyTorch, specifically torch geometric for graph operations. Each model was trained using Adam optimizer with an initial learning rate of $1e^{-3}$ for 200 epochs and a patience of 20, the model with the best accuracy on the validation set was used to report results on the test set.  

\subsection{Ablation study}
\subsubsection{Different mending techniques}
In Table~\ref{tab:table 4} we compare the performance of different mending techniques on the block adjacency matrix. Using an attention-based encoder~\cite{transformer} gives the best accuracy for change detection on the C2D2 dataset. We implemented the LSTM block to model the temporal dynamics in the graphs but the results were not competitive with the encoder. In the modified adjacency matrix $\mathbf{\hat{A}_{B}}$, in addition to temporal connections, the spatial connections are also modified. To evaluate the impact of modified spatial connections, we forced the model to only update the temporal edges, by masking the learned adjacency matrix with zeros in place of spatial edges before adding the skip connection. This resulted in a slight drop in performance which suggests that although the main improvement stems from updating the temporal connections, updating the spatial connections is also beneficial.

\begin{figure}[t]
\begin{center}
\includegraphics[width=0.25\textwidth]{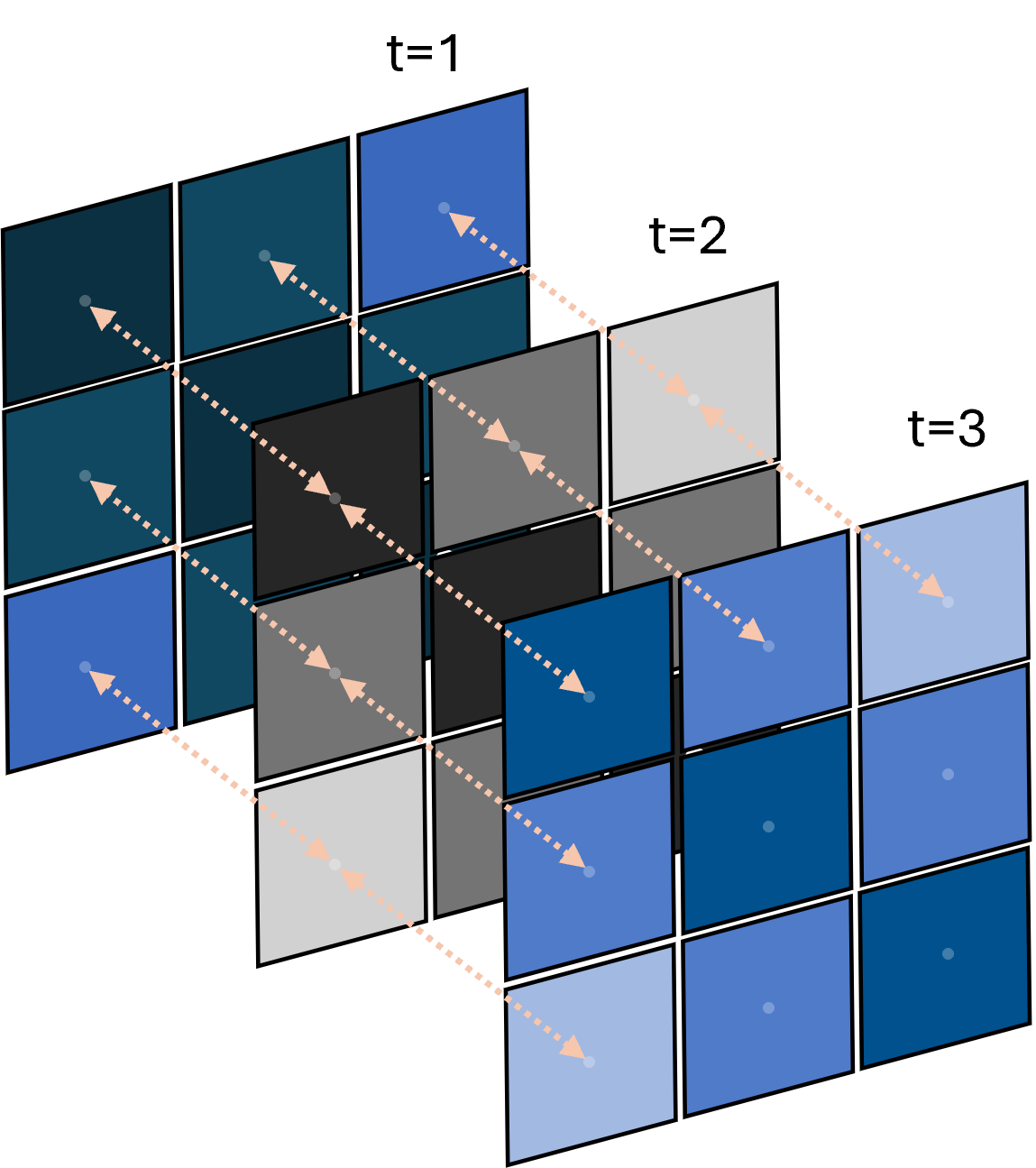}
\end{center}
    \caption{Fixed temporal connection.}
\label{fig:temporalconn}
\end{figure}

Further, we also analyzed three different baseline approaches for mending the adjacency matrix. These included adding fixed temporal connections (see Fig.~\ref{fig:temporalconn}) where the nodes that are in the same spatial location are connected across adjacent time steps, adding random weighted, and random binary edges. Out of these, the fixed $1$ temporal neighbor approach works best with an accuracy of $76.33\%$. We hypothesize that the random edges do not have a significant negative impact on performance because of the use of GAT, which learns to weigh the wrongly added edges appropriately. 
\begin{table}[h] \centering
\caption{Ablative study of different mending techniques on C2D2 dataset.}
\resizebox{0.6\textwidth}{!}{
\begin{tabular}{ccc}
  \toprule
  Method  & \#Param (M)& Accuracy (\%)   \\
  \midrule
  Learnable (encoder)&  $2.98$ & $\mathbf{80.67}$ \\
  Learnable (LSTM) & $0.23$ & $72.67$ \\
  Learnable (temporal edges only)  & $2.98$  & $79.00$ \\
  Fixed 1 temporal neighbour  & $0.05$ & $76.33$ \\
  Random (weighted) & $0.05$ & $75.33$  \\  
  Random (binary) & $0.05$ & $76.00$ \\       
  \bottomrule
\end{tabular}
}
\label{tab:table 4}
\end{table}

\subsubsection{Sparsity}
\begin{figure}[t]
\centering
    \includegraphics[width=0.49\textwidth]{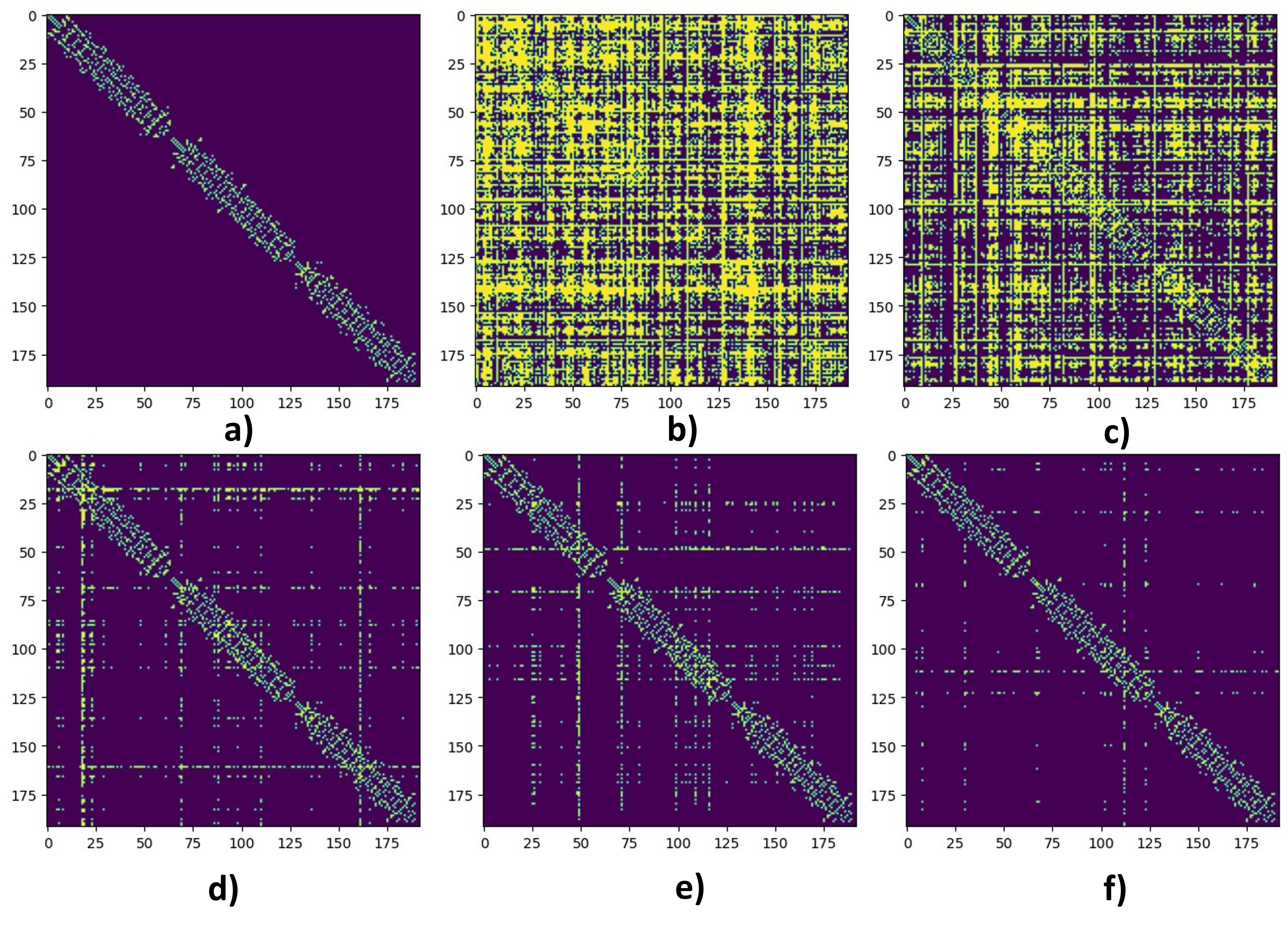}
    \caption{Block adjacency matrix visualization (a) original block adjacency  (b) modified with no norm (c) modified with L2 norm, $\lambda=1e-6$ (d) modified with L1 norm, $\lambda=1e-7$ (e) modified with L1 norm, $\lambda=1e-6$ (f) modified with L1 norm, $\lambda=1e-5$.}
    \label{c2d2_img}
\end{figure}

\begin{table}[b]\centering
\caption{Ablative study on C2D2 Dataset.}
\resizebox{.8\textwidth}{!}{
\begin{tabular}{c|ccc|ccc}
  \toprule
  Method  & Penalty  &Norm & \#Param(M) & Sparsity $\rho$ ($\Delta \rho$) &  $\Delta\mathcal{H}$  & Acc. (\%) \\
  \midrule
    STBAM-64 & $10^{-7}$ & $L1$ & $3.392$ & $0.9396$ ( $0.0347$)&$1.43\times10^0$  & $80.33$\\
    STBAM-64 & $10^{-6}$ & $L1$ & $3.392$& $0.9467$ ($0.0276$)&$4.60\times10^{-1}$  & $\mathbf{80.67}$\\
    STBAM-64 & $10^{-5}$ & $L1$ & $3.392$& $0.9650$ ( $0.0093$)& $6.10\times10^{-1}$ & $80.33$\\
    STBAM-64 & $0$ & $No$ & $3.392$& $0.4779$ ( $0.4964$) &$4.81\times10^4$& $75.00$\\
    STBAM-64 & $10^{-6}$ &$L2$ & $3.392$& $0.6436$ ( $0.3307$) & $1.64\times10^1$ & $77.33$\\
    STBAM-100 & $10^{-6}$&$L1$ & $3.524$ & $0.0042$ ( $0.9801$) & $3.97\times10^2$ & $75.00$\\ 
    \bottomrule
    
\end{tabular}
}
\label{tab:table 1}
\end{table}
We performed the ablative study by comparing $6$ different variants of our method. These variants are generated by changing a number of nodes, the penalty term for sparsity, and the type of norm used. To measure the achieved sparsity $\rho$, we define, for any matrix $\mathbf{S}$ with dimensions $m \times n$,
\begin{equation}
    \rho(\mathbf{S}) = \frac{m \times n - \rm {count}(\mathbf{S} \neq 0)}{m \times n},
\end{equation}
where,
$\rm{count}(\mathbf{S} \neq 0)$ counts the number of non-zero elements in $\mathbf{S}$.
The difference in sparsity between the two block adjacency matrices $\mathbf{A_B}$ and $\mathbf{\hat{A}_B}$ is defined as
\begin{equation}
\Delta \rho=\frac{\sum_{N}\rho(\mathbf{BA})
-\sum_{N}\rho(\mathbf{BA_M})
}{N},
\end{equation}
where the sum over $N$, refers to averaging over the entire test dataset. It can be seen from Table~\ref{tab:table 1} that our best-performing architecture is with $64$ superpixels/image using $L1$ norm and a penalty $\lambda$ of $1 \times 10^{-5}$. Fig.~\ref{c2d2_img} shows the effect of different penalty terms and regularization functions ($L_1$ and $L_2$) on the connectivity of block adjacency matrix. We used this variant of our architecture for comparison with state-of-the-art methods as discussed next.
\subsubsection{Smoothness} 
Data residing on graphs change smoothly across adjacent nodes, which is a well-known assumption for graphical data. The smoothness of a graph given its nodes features $x_1, x_2, ..., x_N \in  \mathbb{R}^d$ and weighted un-directed adjacency matrix $W$
can be quantified by the Dirichlet energy ~\cite{dirichlet_energy}

\begin{equation}
\mathcal{H}=\frac{1}{2}\sum_{i, j} W_{ij}\lVert x_i-x_j \rVert^2=\rm{tr}(X^TLX),
\end{equation} 
\begin{equation*}
    \Delta \mathcal{H}= \mathcal{H}_m-\mathcal{H}_o
\end{equation*}
where $L=D-W$ is the graph laplacian and $D$ is the degree matrix, where $D_{i, i} = \sum_j W_{i, j}$. $\Delta \mathcal{H}$ is the difference in the Dirichlet energy of the graph with the original $\mathcal{H}_o$ and modified $\mathcal{H}_m$ adjacency matrix. In table \ref{tab:table 1} we compare $\Delta \mathcal{H}$ for different models. It has been observed that the smaller $\Delta \mathcal{H}$ leads to better classification accuracy.
\subsection{Comparative Analysis}
We performed comparison with two state-of-the-art algorithms namely 3D-ResNet-34~\cite{c2d2} and STAG-NN-BA-GSP~\cite{1} on C2D2 dataset (Table~\ref{tab:table 2}) and two algorithms by team SK~\cite{surgvisdom} and team Parakeet~\cite{surgvisdom}, and a clip based architecture ~\cite{SDA-clip} on SurgVisdom dataset (Table~\ref{tab:table 3}). It can be seen from Table~\ref{tab:table 2} that our proposed STBAM-64 model achieved the highest accuracy of $80.67\%$ on the C2D2 dataset while having significantly fewer parameters as compared to 3D-ResNet-34. The slight increase in parameters as compared to STAG-NN-BA-GSP is due to the encoder block in our architecture for mending the block adjacency matrix.
\begin{table}[h]\centering
\caption{Comparative  evaluation on C2D2 dataset.}
\resizebox{0.53\textwidth}{!}{
\begin{tabular}{ccc}
  \toprule
  Method  &  \#Param(M) &  Accuracy(\%) \\
  \midrule
  3D-ResNet-34 ~\cite{c2d2} &$63.50$  &  $57.72$    \\
   STAG-NN-BA-E ~\cite{1} &$0.03$  &  $60.02$    \\
  STAG-NN-BA-GCP ~\cite{1} &$0.05$  &  $64.90$    \\
  STAG-NN-BA-GSP ~\cite{1} &$0.03$  &  $77.83$    \\
  STBAM-64-GCN (ours) & $2.98$&$78.57$ \\
  STBAM-64-GAT (ours) &$3.39$  &  $\mathbf{80.67}$    \\
  \bottomrule
  \label{tab:table 2}
\end{tabular}
}
\end{table}

Similarly, on the SurgVisdom dataset, it can be seen from Table~\ref{tab:table 3} that our method achieved the highest F1-score and comparable global F1-score and balanced accuracy compared to other methods from the challenge~\cite{surgvisdom}. We followed the hard domain adaptation task i.e., the training was performed only on VR data, and testing was done on real data. Although, as compared to other methods, we did not introduce any additional steps to solve the domain adaptation problem, our method showed competitive performance. This illustrates the generalization capabilities of our proposed method. The performance of our model is relatively low in comparison to SDA-clip~\cite{SDA-clip}. Our model used $7M$ parameters in comparison to $168M$ parameters used in SDA-clip. The complex architecture of SDA-clip also does not allow for real time inference, which is possible in our method thus allowing it to be used for online applications. 

These results indicate that by addressing the inherent limitations of the block adjacency matrix, which traditionally focuses solely on spatial connections, our encoder block provides a more holistic view of network dynamics by ensuring the graph's connectivity across different time steps, enriching the data representation for more accurate and comprehensive spatiotemporal analysis.


\begin{table}[t] \centering
\caption{Comparative evaluation on SurgVisdom dataset (hard domain adaptation).}
\resizebox{0.8\textwidth}{!}{
\begin{tabular}{cccccc}
  \toprule
 \multirow{2}{*}{Approach}&  \multirow{2}{*}{Method}  & Weighted &Unweighted & Global& Balanced \\
       &  &F1-score & F1-score  & F1-score& Accuracy(\%) \\
  \midrule
  \multirow{6}{*}{Non-graph based} 
  & Rand ~\cite{surgvisdom} &$0.45$&$0.21$& $0.327$  &$32.7$    \\
  & SK ~\cite{surgvisdom}  &$0.46$&$0.22$& $0.370$  &$36.9$    \\
  & Parakeet ~\cite{surgvisdom} & $0.47$&$0.27$& $0.475$ &$55.9$    \\
  & Pure ViT~\cite{SDA-clip}  & $0.53$&$0.39$&$0.536$ &$58.5$    \\
  & ViT+Text~\cite{SDA-clip}  & $0.58$&$0.44$&$0.610$ &$\mathbf{67.7}$    \\
  & SDA-clip~\cite{SDA-clip}  & $\mathbf{0.66}$&$\mathbf{0.55}$&$\mathbf{0.647}$ &$60.9$    \\    
  \midrule
  \multirow{2}{*}{Graph-based} 
  & GAT-BA  & $0.44$&$0.34$& $0.460$ &$49.3$    \\
  & STBAM-64 (ours)  & $0.54$&$0.40$& $0.561$ &$44.7$    \\
    
  \bottomrule
\end{tabular}
}
\label{tab:table 3}
\end{table}


\subsection{Eigenvalue and Fiedler Value Analysis}
\begin{figure}[t]
  \centering
  \subfigure[]{
    \includegraphics[width=0.5\linewidth]{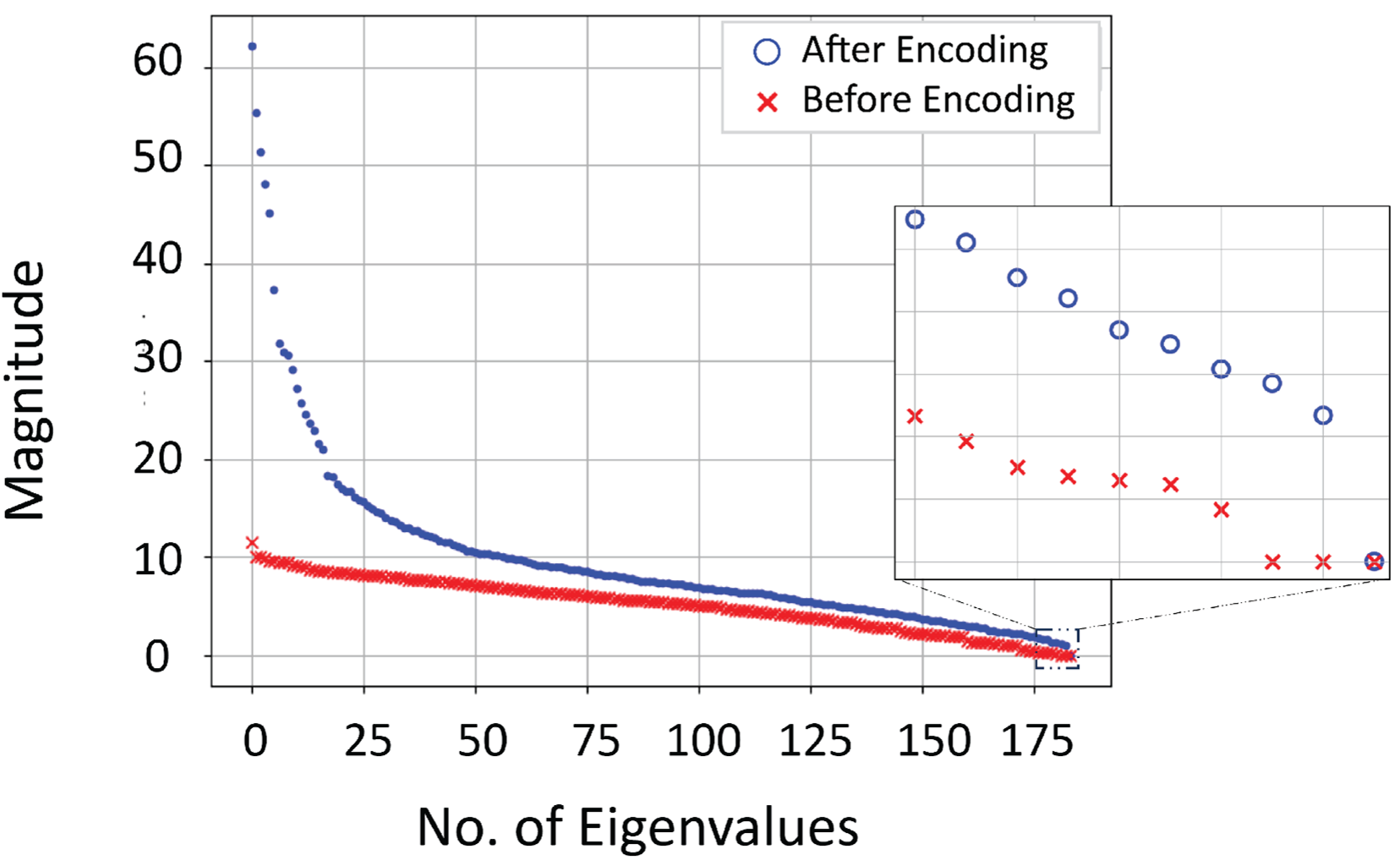}
    \label{fig:subfig1}
  }
  \subfigure[]{
    \includegraphics[width=0.4\linewidth]{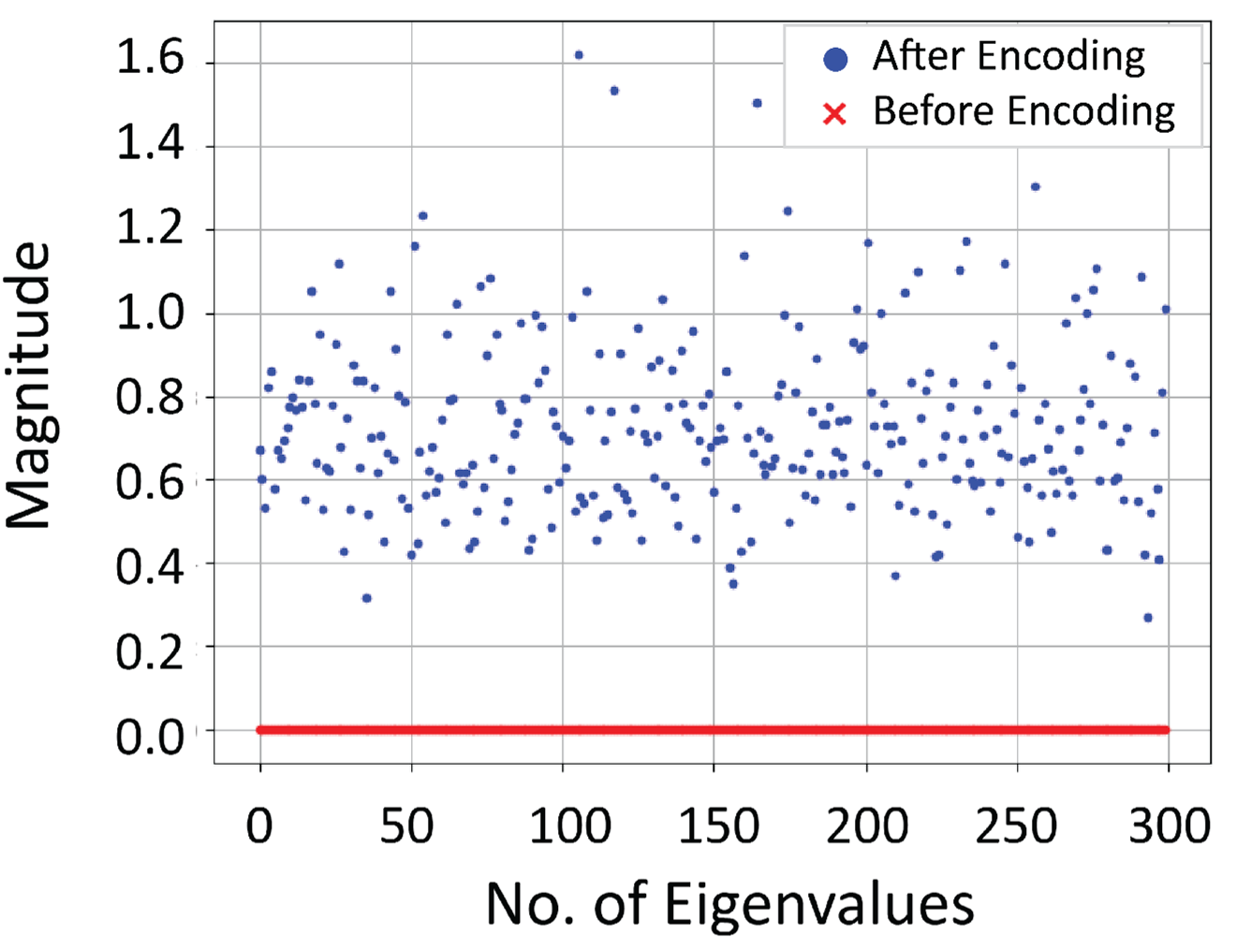}
    \label{fig:subfig2}
  }
  \caption{Interpretation of graph connectivity using eigenvalues a) Absolute eigenvalues of Laplacian matrix before $\&$ after encoding b) Fiedler values on the test set before $\&$ after encoding.}
  \label{fig:4}
\end{figure}

The number of zero eigenvalues in the Laplacian matrix can be insightful regarding graph connectivity, in Fig.~\ref{fig:4}(a) we plot the eigenvalues of the block adjacency matrix before and after encoding for a test sample from the C2D2 dataset. Specifically, N zero eigenvalues indicate the presence of N unconnected sub-graphs. In our original matrix, we observe 3 zero eigenvalues, corresponding to 3 time steps of data, suggesting a fragmented graph. In contrast, our modified matrix has only 1 zero eigenvalue for 1 fully connected graph, further, we see the trend that eigenvalues of the modified matrix are generally greater in magnitude indicating improved graph connectivity, which aligns with the successful integration of temporal information by the encoder block.

Figure~\ref{fig:4}(b) illustrates the Fiedler values (also known as the Algebraic Connectivity of the graph) of the Laplacian of the block adjacency matrix before and after the encoding block for the entire test set of the C2D2 dataset. The Fiedler value, which is the second smallest eigenvalue of the Laplacian matrix, serves as an indicator of a graph's connectivity~\cite{Fiedler1973}. A Fiedler value of zero generally suggests that the graph is disconnected or poorly connected. In our analysis, the Fiedler values of the original block adjacency matrix are all zero, underlining the presence of disconnected sub-graphs. On the other hand, the modified block adjacency matrices show, on average, much higher Fiedler values. This indicates that our proposed model while addressing the issue of unconnected sub-graphs enhances the overall graph connectivity, thereby facilitating more effective message passing in the spatio-temporal graph.

\section{Conclusion}
\label{conclusion}
This work introduces a significant contribution by extending graph convolution message-passing in the block adjacency matrix to incorporate temporal dimensions. Traditional models, constrained by the block adjacency matrix, limit message-passing to spatial nodes, hindering the capture of dynamic changes over time. The introduced encoder block effectively bridges this gap, enabling a more nuanced understanding of spatio-temporal data and enhancing predictive capabilities. We develop an effective architecture that allows a learnable representation of spatio-temporal data as graphs and adeptly employs a single GNN to simultaneously capture both spatial and temporal aspects of data facilitating the learning of comprehensive representations.

Future works include extending the block adjacency matrix representation to different spatio-temporal graph-based problems e.g., social network analysis, traffic management, and bio-informatics. Further, better features for the super-pixels can be explored e.g., features from remote-clip can be used when working with satellite image time series data.

\bibliographystyle{splncs04}
\bibliography{strings}

\begin{thebibliography}{10}
\providecommand{\url}[1]{\texttt{#1}}
\providecommand{\urlprefix}{URL }
\providecommand{\doi}[1]{https://doi.org/#1}

\bibitem{EvolveGCN}
A, P., G, D., J, C., T, M., T, S., H, K., T, K., T, S., C, L.: Evolvegcn: Evolving graph convolutional networks for dynamic graphs. InProceedings of the AAAI conference on artificial intelligence pp. 5363--5370 (2020)

\bibitem{surgvisdom}
A, Z., et~al.: Surgical visual domain adaptation: Results from the miccai 2020 surgvisdom challenge. arXiv preprint arXiv:2102.13644  (2021)

\bibitem{TCN}
Bai, S., Kolter, J.Z., Koltun, V.: An empirical evaluation of generic convolutional and recurrent networks for sequence modeling. arXiv:1803.01271  (2018)

\bibitem{dirichlet_energy}
Belkin, M., Niyogi, P.: Laplacian eigenmaps and spectral techniques for embedding and clustering. in Neural Information Processing Systems (NIPS)  (2001)

\bibitem{c2d2}
Bhimra, M.A., Nazir, U., Taj, M.: Using 3d residual network for spatio-temporal analysis of remote sensing data. ICASSP pp. 1403--1407 (2019)

\bibitem{yield_pred}
Bose, P., Kasabov, N.K., Bruzzone, L., Hartono, R.N.: Spiking neural networks for crop yield estimation based on spatiotemporal analysis of image time series. in IEEE Transactions on Geoscience and Remote Sensing  \textbf{54}(11),  6563--6573 (2016)

\bibitem{4}
Censi, A.M., et~al.: Attentive spatial temporal graph cnn for land cover mapping from multi-temporal remote sensing data. IEEE Access  \textbf{9},  23070 -- 23082 (2021)

\bibitem{flood_forecast}
Ding, Y., Zhu, Y., Feng, J., Zhang, P., Cheng, Z.: Interpretable spatio-temporal attention lstm model for flood forecasting. Neurocomputing  \textbf{403},  348--359 (2020)

\bibitem{Fiedler1973}
Fiedler, M.: Algebraic connectivity of graphs. Czechoslovak Mathematical Journal  \textbf{23}(2),  298--305 (1973), \url{http://eudml.org/doc/12723}

\bibitem{ride_hailing}
Geng, X., et~al.: Spatiotemporal multi-graph convolution network for ride-hailing demand forecasting. in Proceedings of the AAAI conference on artificial intelligence  \textbf{33}(1),  3656--3663 (2019)

\bibitem{modular_graph_transformer}
HD, N., XS, V., DT, L.: Modular graph transformer networks for multi-label image classification. InProceedings of the AAAI conference on artificial intelligence  \textbf{35}(10),  9092--9100 (2021)

\bibitem{feddy}
Jiang, M., Jung, T., et~al.: Federated dynamic graph neural networks with secure aggregation for video-based distributed surveillance. ACM Transactions on Intelligent Systems and Technology  \textbf{13}(4),  1--23 (2022), \url{https://doi.org/10.1145/3501808}

\bibitem{STGNN}
Jin, G., Liang, Y., Fang, Y., Huang, J., Zhang, J., Zheng, Y.: Spatio-temporal graph neural networks for predictive learning in urban computing: A survey (2023)

\bibitem{euler}
King, I.J., Huang, H.H.: Euler: Detecting network lateral movement via scalable temporal link prediction. ACM Transactions on Privacy and Security  (2023)

\bibitem{GCNconv}
Kipf, N., T., Welling, M.: Semi-supervised classification with graph convolutional networks. arXiv preprint arXiv:1609.02907.  (2016)

\bibitem{1}
Nazir, U., Islam, W., Taj, M., Khalid, S.: Spatio-temporal driven attention graph neural network with block adjacency matrix (stag-nn-ba). AAAI Symposium  (2023)

\bibitem{RNN}
Rumelhart, D.E., Hinton, G.E., Williams, R.J.: Learning representations by back-propagating errors. Nature pp. 533--536 (1986)

\bibitem{VGG16}
Simonyan, K., Zisserman, A.: Very deep convolutional networks for large-scale image recognition. arXiv:1409.1556  (2014)

\bibitem{land_cover}
Vali, A., Comai, S., Matteucci, M.: Deep learning for land use and land cover classification based on hyperspectral and multispectral earth observation data: A review. Remote Sensing  \textbf{12}(15), ~2495 (2020)

\bibitem{transformer}
Vaswani, A., et~al.: Attention is all you need. Advances in neural information processing systems  (2017)

\bibitem{Graph_attention_networks}
Velickovic, Petar, Cucurull, G., Casanova, A., Romero, A., Lio, P., Bengio, Y.: Graph attention networks. ICLR  (2018)

\bibitem{ST_data}
Wang, S., Cao, J., Yu, P.S.: Deep learning for spatio-temporal data mining: A survey. in IEEE Transactions on Knowledge and Data Engineering  \textbf{34}(8),  3681--3700 (2022)

\bibitem{video_action}
Wu, D., Sharma, N., Blumenstein, M.: Recent advances in video-based human action recognition using deep learning: A review. International Joint Conference on Neural Networks (IJCNN) pp. 2865--2872 (2017)

\bibitem{GNN_survey}
Wu, Z., et~al.: A comprehensive survey on graph neural networks. IEEE Transactions on Neural Networks and Learning Systems  \textbf{32}(1),  4--24 (2021)

\bibitem{splitting_algorithm}
Yamada, K., Tanaka, Y., Ortega, A.: Time-varying graph learning based on sparseness of temporal variation. In ICASSP  (2019)

\bibitem{traffic_forecasting}
Yu, B., Yin, H., Zhu, Z.: Spatio-temporal graph convolutional networks: A deep learning framework for traffic forecasting. arXiv preprint arXiv:1709.04875  (2017)

\bibitem{STAR}
Yu, C., et~al.: Spatio-temporal graph transformer networks for pedestrian trajectory prediction. ECCV, Glasgow, UK, August 23–28, 2020, Proceedings pp. 507--523 (2020)

\bibitem{SDA-clip}
Yuchong, L., et~al.: Sda-clip: surgical visual domain adaptation using video and text labels. Quantitative Imaging in Medicine and Surgery  (2023)

\bibitem{GTN}
Yun, S., Jeong, M., Kim, R., Kang, J., Kim, H.J.: Graph transformer networks. Advances in neural information processing systems  (2019)

\bibitem{eeg}
Zhang, D., Yao, L., Chen, K., S.~Wang, X.C., Liu, Y.: Making sense of spatio-temporal preserving representations for eeg-based human intention recognition. in IEEE Transactions on Cybernetics  \textbf{50}(7),  3033--3044 (2020)

\end{thebibliography}

\end{document}